# Table Tennis Stroke Detection and Recognition Using Ball Trajectory Data

Kaustubh Milind Kulkarni
kmkapril15@gmail.com

Rohan S Jamadagni
rohanjamadagni@gmail.com

Jeffrey Aaron Paul
jeffrey.paul2000@gmail.com

Sucheth Shenoy
sucheth17@gmail.com

## Abstract

*In this work, the novel task of detecting and classifying table tennis strokes solely using the ball trajectory has been explored. A single camera setup positioned in the umpire's view has been employed to procure a dataset consisting of six stroke classes executed by four professional table tennis players. Ball tracking using YOLOv4, a traditional object detection model, and TrackNetv2, a temporal heatmap based model, have been implemented on our dataset and their performances have been benchmarked. A mathematical approach developed to extract temporal boundaries of strokes using the ball trajectory data yielded a total of 2023 valid strokes in our dataset, while also detecting services and missed strokes successfully. The temporal convolutional network developed performed stroke recognition on completely unseen data with an accuracy of 87.155%.*

*Several machine learning and deep learning based model architectures have been trained for stroke recognition using ball trajectory input and benchmarked based on their performances. While stroke recognition in the field of table tennis has been extensively explored based on human action recognition using video data focused on the player's actions, the use of ball trajectory data for the same is an unexplored characteristic of the sport. Hence, the motivation behind the work is to demonstrate that meaningful inferences such as stroke detection and recognition can be drawn using minimal input information.*

## 1. Introduction

Ball tracking for sports is a popular task in computer vision with a wide range of practical applications. Data from ball tracking can be used for automated umpiring systems, generate game analytics as well as provide technical feedback to the players, and aid in sports training. Table tennis is one of the fastest racket sports involving a range of different strokes. Presently, table tennis ball tracking systems are limited to umpiring purposes and kinematic parameters of the ball during gameplay in order to increase the popularity and public outreach of the sport.

However, the instances of ball trajectory data being used to bring about details regarding the technical aspects in table tennis such as the strokes executed are scarce. Hence, we present our work carried out in this direction, which involves detecting and recognizing strokes executed by professional table tennis players using two-dimensional ball trajectory data obtained from a single camera feed.

Table tennis strokes vary from each other in terms of speed and spin imparted on the ball. On manual inspection, slow strokes such as pushes and lobs can be easily distinguished from fast strokes like flicks and flats based on the ball velocity along the length of the table. The presence of spin on the ball in the case of topspin stroke results in a Magnus force acting in the downward direction on the ball and hence, affecting the trajectory of the ball. Therefore, although being of similar speeds, the strokes involving spin on the ball like topspin strokes have discernible differences in the trajectory of the ball as compared to the block strokes, which lack the same amount of spin. The work presented in this paper builds upon these premises and thus, explores the possibility of performing stroke recognition for table tennis using only the two-dimensional ball trajectory data.

## 2. Related works

The preliminary works involving ball tracking in table tennis mainly focused on developing fast models which track the ball to aid an umpire's judgments in the sport. For example, the work carried out by Wong and Dooley [1] discusses algorithms exploiting the spatial and temporal information from table tennis match videos to detect the ball during the toss of a service and provides its decision on the validity of the service based on the rules of the game. Color segmentation and stable principal component pursuit methods have been used by Chang-Hung Hung [2] for detecting the ball present in the frames captured by a high-speed camera during ball toss of a service and hence identify illegal ball toss based on the ball trajectory. Hnin Myint et al. [3] have used stereo videos captured by a system of two cameras to detect and track a table tennis ball and generate the three-dimensional trajectory of the ball, developing a precursor for automatic umpiring systems.

The more recent works in the domain have been focussed on the implementation of machine learning and deep learning based approaches for table tennis ball tracking and analysis of the ball trajectory. Yun-Feng Ji et al. [4] have implemented a Visual Object Detection with a Computational attention System (VOCUS) system for image segmentation using color channel information and have used machine learning algorithms to detect table tennis balls in the form of a smear as captured by the frames of a low-speed camera, with the best accuracy of 87%. A method to obtain three-dimensional kinematic parameters of the table tennis ball using a single camera feed is presented by Jordan Calandre et al. [5][6], where a 2D convolutional neural network is used to extract the apparent ball size and hence, derive the depth information for the ball position. Fufeng Qiao [7] has discussed the application of deep learning based Deep Convolutional Neural Network-Long Short Term Memory (DCNN-LSTM) model for ball tracking and prediction of the ball trajectory.

Apart from works based on computer vision, Ralf Schneider et al. [8] have conducted a statistical analysis on a database of table tennis ball trajectories and have discussed the effects of ball size, weight, spin, and speed on the trajectory.

The task of stroke detection and classification is an emerging application of computer vision and pattern recognition in table tennis. Though plenty of research work has been carried out using sensor-based approaches, only a reasonable amount of work has been carried out in this purview solely based on video data, focusing on action recognition performed on the player. A combination of vision and IMU sensor-based approach has been developed by Yapeng Gao et al. [9] for stroke recognition using 6D racket pose data. Stroke recognition and detection with an average accuracy of 88% - 100% for the forehand and backhand variations of the push stroke has been carried out using three-dimensional pose estimation data from IR depth camera by Habiba Hegazy et al. [10]. Kadir Atkas et al. [11] have implemented a spatiotemporal combination of RGB and optical flow based methods for stroke recognition task, obtaining a test accuracy 90.7% on the MediaEval Challenge 2020 dataset. The application of optical flow singularity for stroke detection has been discussed by Jordan Calandre [12]. They obtained a best accuracy of 14.12% on the test set of the MediaEval 2019 Sports Video Annotation "Detection of Strokes in Table Tennis" task [13]. A twin spatiotemporal convolutional network implementing 3D attention mechanism has been used by Pierre-Etienne Martin et al. [14] for stroke classification. The attention blocks show a 5% increase in the classification accuracies compared to their baseline models. Pierre-Etienne Martine et al. [15] have also introduced a three-stream (RGB, optical flow, and pose estimation) 3D/1D CNN model for stroke classification and detection tasks.

SPIN, a multi-task dataset for tracking and action recognition in table tennis, has been presented by Steven Schwarcz et al. [16]. They have also discussed tasks such as ball tracking, spin prediction, and pose detection on their dataset. TTNet, a network for real-time temporal and spatial event spotting for table tennis has been introduced by Roman Voeikov et al. [17]. The network provided a 97.0% test accuracy in the event spotting task on their dataset named OpenTTGames. KM Kulkarni and S Shenoy [18] have described an efficient method for collecting stroke video dataset and have used 2D pose estimation data for stroke recognition. They obtained the best validation accuracy of 97.37% and a generalization accuracy of 98.72% using a Temporal Convolutional Network (TCN) model.

Although an abundant amount of work has been carried out in the field of table tennis with respect to ball detection and tracking, and stroke detection and recognition individually, there is a lack of literature addressing a combination of the two.

## 3. Dataset collection

Although there are several open-source datasets available for table tennis along with broadcast footage that can be used for the tasks in our work, none of them provide a standardized view of the ball trajectory during gameplay. There are also issues such as ball occlusion due to movements of the players leading to the discontinuous appearance of the ball in the frames, hiding parts of the trajectory of the ball. Though there exists broadcast footage from desirable viewing angles, the low frame rates at which they are captured do not facilitate the level of accuracy we require for ball tracking due to significant distortion in the shape of the ball. A distinct view where the bounce of the ball is evident is an important part of our process, and camera angles with which the majority of broadcast videos are captured do not encourage this. Therefore, there was a need to collect a custom dataset that enabled an occlusion-free view of the ball trajectory, while also simplifying the labeling process of the dataset.

The umpire's view or the side view of the table provides an optimal viewing angle to observe the trajectory of the ball, aiding the extraction of useful features. It enables stroke detection from the ball trajectory based on changes in the horizontal component of the ball position. An elevation in the camera position above the table also facilitates the extraction of the ball pitch position on the table using perspective transformation.

### 3.1. Setup

A GoPro Hero 8 Black camera, is placed at a distance of 1750 mm from the edge of the table along the axis of the net and at an elevation of 1400 mm from the ground level. The entire table is captured in the frame as shown in Fig. 1 when the camera is positioned as depicted in Fig. 2. The camera angle was set such that both ends of the table were in the field of view of the camera. The videos were recorded at a resolution of 1920x1080 pixels, and a frame rate of 120 frames per second (fps) in order to better capture the fast-moving ball.

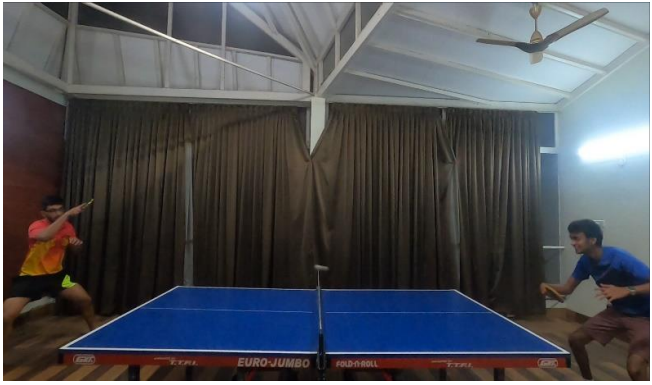

Figure 1: A frame captured by our camera setup.

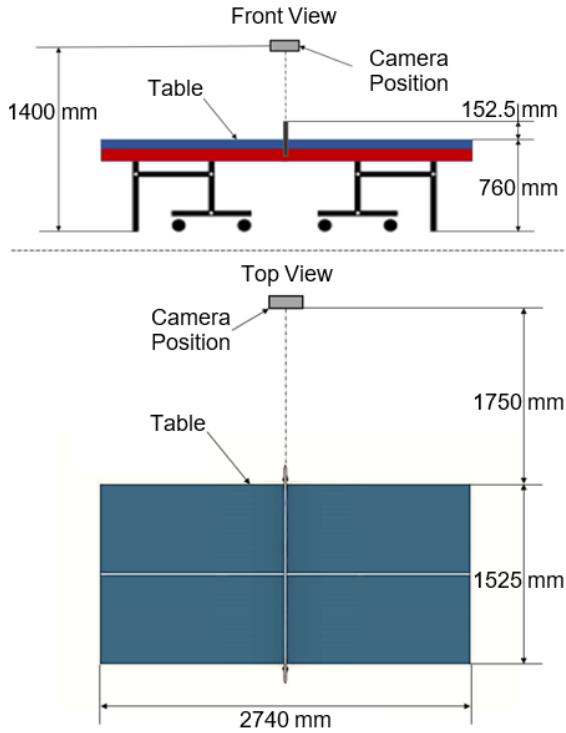

Figure 2: Camera placement for dataset generation.

The strokes considered in our dataset include topspins, blocks, pushes, flicks, lobs, and flats. Both forehand and backhand versions of all these strokes were executed by the players in the dataset along with variations in the ball placement such as cross-court (along the diagonal of the table), down-the-line (along the length of the table), center-to-center (along the centerline), and all other variations combining the above. Data was collected from the strokes executed by four professional table tennis players, out of which three were used in the training dataset and one was considered exclusively for the test dataset. The statistics of our dataset are as shown in Tab. 1.

|   | Stroke Type | Number of strokes in training dataset | Number of strokes in validation dataset |
|---|---|---|---|
| 1 | Topspin | 262 | 80 |
| 2 | Block | 260 | 101 |
| 3 | Push | 236 | 127 |
| 4 | Flick | 232 | 74 |
| 5 | Lob | 225 | 94 |
| 6 | Flat | 263 | 69 |
|   | Total | 1478 | 545 |

Table 1: Stroke-wise distribution of the dataset.

## 4. Ball tracking

It is essential to perform ball tracking with high accuracy since this data has been leveraged to extract information about a stroke. In table tennis, the ball can move at speeds greater than 100 km/h. Even while recording at high frame rates, there is some distortion in the image (as shown in Fig. 1) of the ball when fast strokes are executed. The varying appearance of the ball and lack of physical features further add to the complexity of this task. Hence, accurate tracking of the ball is a challenging task.

The objective of this section is to detect the ball and obtain its two-dimensional coordinates in each frame of our dataset, where the top-left corner of the frame is defined as the origin of the pixel space along with x and y axes in the horizontal and vertical directions respectively. This is performed using two distinct deep-learning approaches - object detection using YOLOv4, and TrackNetv2 which uses a neural network to output a heatmap of the location of the ball in a given frame.

### 4.1. YOLOv4

YOLOv4 is state-of-the-art in terms of object detection algorithms and performs well with regard to detecting the objects in a given frame. The object detection algorithm makes use of the Darknet53 network trained on MS COCO for feature extraction [19]. The YOLOv4 model was trained on 68,467 images of badminton matches where the shuttlecock was labeled [20], followed by 968 images of table tennis matches where the ball was labeled in various conditions. The badminton data was used for training in addition to the table tennis data since the two sports are very similar in the aspect that the object to be tracked is of similar size and color, traveling at similar speeds and range of motion in a frame. YOLOv4 is promising when it comes to the detection of objects in single images but fails to consider the relative position of the ball across the temporal dimension and proceeds to make a new prediction for every single input frame. Although accurate, this could lead to several false positives due to balls in the background or static objects having an appearance similar to the ball.

### 4.2. TrackNetv2

TrackNetv2 [20] was developed to track tiny, fast-moving objects in sports applications. Instead of detecting the bounding boxes of the ball, this model proceeds to generate a heatmap of the probable locations of the ball in a given frame. The network is designed to handle the occlusions of the ball since it takes multiple frames as the input. This enables the model to use temporal information gathered to predict the ball coordinates in each frame. The TrackNetv2 model was trained on the same badminton and table tennis data as the YOLOv4 model.

### 4.3. Results

In order to compare the aforementioned methods, the ground truth for 2 different videos was labeled. The first video involves a professional player playing push strokes at a medium to slow pace, and in the second, the same player executes continuous topspin strokes with the ball traveling at high velocity. The two scenarios provide a clear distinction between the type of strokes that are executed in a match by professional players and the results obtained from both approaches can be accurately measured and benchmarked.

We define the conditions for True Positive and False Positive cases as follows: (a) A set of predicted coordinates $(\bar{x}, \bar{y})$ is considered to be a True Positive if the Euclidean distance measured between the predicted point and the ground truth of the ball $(x, y)$ is less than or equal to N=25 pixels. This threshold of N is set because the average diameter of the ball is 30 pixels when the video's resolution is set at 1920x1080. (b) The prediction is considered to be a False Positive if the condition fails or if there is a positive prediction for a negatively labeled frame. We also define the conditions for True Negative and False Negative cases as follows: (a) The prediction is considered to be a True Negative if there is a negative prediction for a negatively labeled frame. (b) A set of predicted coordinates $(\bar{x}, \bar{y})$ is considered to be a False Negative if the condition fails or if there is a negative prediction for a positively labeled frame. The equation for the Euclidean distance is given by (1).

$$d = \sqrt{(\bar{x} - x)^2 + (\bar{y} - y)^2} \quad (1)$$

TrackNetv2 makes a single coordinate as a prediction, i.e. (x, y) coordinate in pixel space which represents the center of the ball, while YOLOv4 on the other hand, predicts the center along with the height and width of the bounding box surrounding the ball. In order to benchmark the metrics between the two approaches, only the predicted center has been considered.

Although the TrackNetv2 model considers the relative position of the ball with the aid of the temporal aspect of the data, the F1-scores from Tab. 2 reveal that YOLOv4 outperforms TrackNetv2 in both cases - slow and fast ball velocity datasets. This is mainly due to the higher number of false negatives predicted by the TrackNetv2 model. The inference speed for TrackNetv2 is 20 frames per second whereas that of YOLOv4 is 60 frames per second when run on an Nvidia Geforce RTX 3070 GPU. Hence, YOLOv4 is used to detect the ball position for all the videos in our dataset and proceed further with the stroke detection and recognition tasks.

|  | Fast (Topspin) | | Slow (Push) | |
| --- | --- | --- | --- | --- |
| Model | YOLOv4 | TrackNetv2 | YOLOv4 | TrackNetv2 |
| True Positive | 374 | 369 | 396 | 390 |
| True Negative | 290 | 281 | 289 | 280 |
| False Positive | 18 | 17 | 5 | 12 |
| False Negative | 11 | 26 | 0 | 8 |
| Precision | 0.9541 | 0.9560 | 0.9875 | 0.9701 |
| Accuracy | 95.82 | 93.80 | 99.28 | 97.10 |
| F1 Score | 0.9627 | 0.9449 | 0.9937 | 0.9750 |

Table 2: Ball tracking models benchmarked for fast and slow ball velocity data.

## 5. Stroke detection

As discussed in Section 4, YOLOv4 was used to extract the ball coordinates in all the raw untrimmed videos of our dataset. To automate the process of splitting individual strokes from the data-frame, we use a purely mathematical approach on these extracted coordinates. As elucidated in this section, the temporal boundaries have been determined for each stroke executed by the players in each video of the dataset. This information is further used to train and validate the various stroke recognition models developed in Section 6.

|  | Detection Class | Overall Accuracy (%) |
| --- | --- | --- |
| 1 | Total Strokes | 95.327 |
| 2 | Serves | 100.00 |
| 3 | Missed Strokes: Net | 92.857 |
| 4 | Missed Strokes: Out | 90.900 |
| 5 | Valid Strokes | 93.850 |

Table 3: Overall accuracies obtained for the various classes of detection task over the entire dataset.

Tab. 3 showcases the overall accuracies obtained by the mathematical approach developed for stroke detection over the entire dataset collected. The accuracies have been evaluated by comparing the number of strokes detected for the various classes with the number of strokes obtained for those classes via manual inspection.

### 5.1. Data pre-processing

The ball tracking model's output data-frame consists of the x and y coordinates of the ball for each frame in the untrimmed video, obtained from the trained YOLOv4 model. The top left corner of the frame is considered to be the origin of the Cartesian coordinate system for ball detection. The rows of the data-frame corresponding to the timestamps where the ball is not detected are dropped. An example of the curves for the ball's x and y coordinates thus obtained are depicted in Fig. 3.

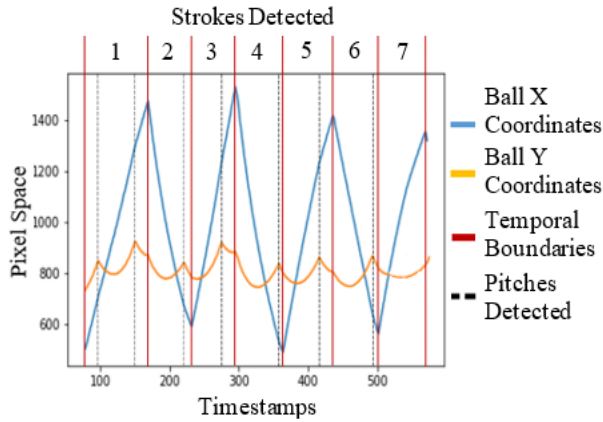

Figure 3: Curves for Ball X and Y coordinates after data pre-processing.

As the ball travels from one side of the table to the other during a stroke, the x coordinate of the ball detected x(t) varies from a local minima to a local maxima or vice-versa depending upon the side from which the stroke has been executed. Hence, temporal boundaries for each stroke as well as the direction of the stroke (either left-to-right or right-to-left) are determined by extracting the timestamps corresponding to each pair of consecutive local extrema detected on the temporal curve of the x coordinates of the ball. A temporal sliding window (a,b) of length 83.33 milliseconds (10 frames) was used for the extraction of the temporal boundaries of the strokes using (2). The temporal boundaries for the strokes obtained are depicted by the red vertical lines in Fig 3. These strokes detected include valid strokes, missed strokes as well as serves.

$$(x, y, t)_{End\ of\ right-to-left\ Stroke} = (x(t), y(t), t)$$
$$: \exists\ (a, b): a < t < b\ \&\ x(t) \leq x(z) \forall z \in (a, b)$$

$$(x, y, t)_{End\ of\ left-to-right\ Stroke} = (x(t), y(t), t)$$
$$: \exists\ (a, b): a < t < b\ \&\ x(t) \geq x(z) \forall z \in (a, b)$$
(2)

### 5.2. Valid strokes and pitch detection

In table tennis, a valid stroke is one in which the ball traverses from the player's side of the table to the opponent's side, while pitching at least once on the opponent's side of the table. In the scope of our dataset, there are no strokes where the ball pitches more than once on the opponent's side of the table. The ball pitch on the table produces a local maxima in the curve of the y coordinates of the ball as illustrated by the dashed black lines in Fig. 3. A temporal sliding window (a,b) of length 41.67 milliseconds (5 frames) was used for the detection of ball pitch events using (3). The valid strokes are detected with the condition that there is only one ball pitch occurring between the temporal boundaries of the stroke.

$$(x, y, t)_{Ball\ Pitch} = (x(t), y(t), t): \exists\ (a, b): a < t < b\ \&\ y(t) \geq y(z) \forall z \in (a, b)$$
(3)

In a few cases where the ball is occluded by the racket at either the beginning or the end of the stroke, a local maxima is seen at this timestamp as observed at the end of strokes numbered 1 and 3 in Fig 3. Such maxima in the ball's y coordinate curve occurring close to the temporal boundaries of the strokes are not considered pitches.

### 5.3. Missed strokes detection

A missed stroke in table tennis is either a stroke that is played into the net or a stroke where the ball fails to pitch on the opponent's side of the table. A stroke played into the net has a local extrema in the ball's x coordinate curve occurring at lower values of x as the ball does not traverse the entire length of the table. Therefore, such strokes are detected with a threshold for x coordinates at the temporal boundaries of the strokes, roughly corresponding to the position of the net along the X-axis. A stroke with no pitches detected within its temporal boundaries (stroke number 7 in Fig. 3) is classified as a missed stroke of the latter type.

### 5.4. Service detection

The rule for services in the sport dictates the requirement of two pitches, one on each side of the table for the validity of the serve. Hence, strokes with two pitches detected within the temporal boundaries are classified as services. In Fig. 3, stroke number 1 is thus classified as a serve.

# 6. Stroke recognition

One of the main aspects of a table tennis player is the player's stroke. Recognizing a stroke is a stepping stone towards analyzing it in detail and providing technical insights to a player. Each distinct stroke has unique characteristics with respect to the ball. The variation in the speed and trajectory of the ball provides an insight into which stroke has been played, without the need to observe the player playing the stroke. Ball trajectory data for each stroke was obtained using the methods mentioned in Section 4 and Section 5.

In a rally, the ball's trajectory varies from left-to-right and right-to-left along the pixel space of the frame, as two players play back and forth. As a standardized method to input data, ball trajectory data was fed to the model in a left-to-right fashion. Using the stroke detection algorithm developed (Section 5), the direction from which the ball was played can be distinguished. Hence, to account for the data where the ball moves from the right to the left, the origin was shifted from the top-left corner to the top-right corner, essentially mirroring the ball trajectory along the y axis.

Different speeds of each stroke yield varying amounts of time steps in the data. For input to the model, we chose 200 time steps (approximately 1.67 seconds at 120 fps) as our standard input time step size. Each time step had 2 features, the x and y coordinates of the ball. In the event that the input time series data had fewer than 100 time steps, the input sequence was padded with zeros. Both, pre-padding and post-padding approaches have been implemented and compared. This data was fed into various stroke recognition models which then classified it into one of 6 pre-defined strokes.

Sections 6.1 and 6.2 elaborate on the different methods used to recognize a stroke just based on ball trajectory data. For each of these methods, a training-validation split of 85%-15% has been used, and the strokes of a separate player was used for the test dataset

## 6.1. Machine learning approach

To input the data into a machine learning model, the data was first flattened to 400 features per sample (200x2 where 200 is the number of time steps per sample and 2 is the number of features per time step i.e. x and y coordinates of the ball). This flattening of the data results in the elimination of the temporal dimension of the data. Various machine learning models were used to fit this data and the best results obtained after tuning the hyperparameters are as shown in Tab. 4.

|  | Pre-Padded | | Post-Padded | |
| --- | --- | --- | --- | --- |
| Model | Validation (%) | Test (%) | Validation (%) | Test (%) |
| Support Vector Machine (RBF kernel, C=10) | 89.473 | 77.798 | 90.460 | **82.935** |
| Random Forest Classifier (n_est=25) | **91.776** | 68.990 | **93.092** | 76.880 |
| XGBoost - Random Forest | 89.802 | 68.073 | 92.434 | 76.697 |
| XGBoost | 90.789 | 66.055 | 92.434 | 77.064 |
| K-Nearest Neighbors (n=9) | 82.565 | **82.385** | 75.328 | 76.513 |

Table 4: Comparison of machine learning models.

When a Random Forest classifier was used to classify the strokes, the best test results were obtained when the number of trees in the model was limited to 25, and the input was post-padded with zeros. We also trained a multi-class Support Vector Machine (SVM) using a Radial Basis Function (RBF) as the kernel for classification. Although the variance was high, the SVM model with post-padded inputs provided the best test accuracy, using a one-vs-rest approach, and the normalization constant at the value c=10. An XGBoost model with a Random Forest classifier generalized better over the test data than an XGBoost classifier without using Random Forest when the input was pre-padded. On the contrary, it was observed that when the input was post-padded, the XGBoost classifier outperformed the XGBoost model with a Random Forest Classifier in the task of generalizing over the test set. For the K-Nearest Neighbors model, we set the number of nearest neighbors k=9. As the value of k decreased, it was observed that the validation accuracy increased but the test accuracy dropped, and as the k value was increased, both the accuracies saw decremental changes.

As seen from Tab. 4, it can be concluded that the KNN algorithm was the best out of all the machine learning models for the stroke recognition task using pre-padded inputs, since the accuracies that were obtained, although indicative of high bias, exhibit low variance. As for the models trained on post-padded inputs, it was observed that the developed SVM model outperformed all other models

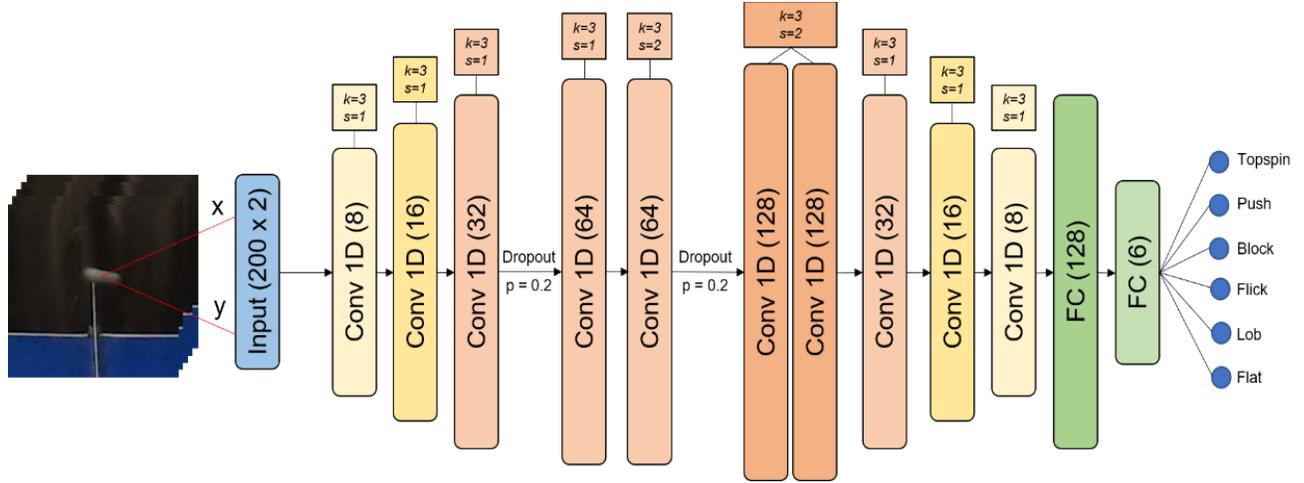

Figure 4: TCN model architecture. k – kernel size, s – stride.

on the test set, with the highest accuracy across all variations of the models.

## 6.2. Deep learning approach

In our deep learning approach, the temporal dimension of the input data was kept intact. The 200-frame input consisting of x and y coordinates was fed into a neural network model that then classified the input stroke into one of six pre-defined classes. For time series classification, we developed three architectures - An LSTM model [21], a TCN model [22], and a purely dense, fully connected model (FCNN - Fully Connected Neural Network). This main architecture was followed by a classifier block consisting of 2 fully connected layers and maintained throughout all architectures. In these last two layers, the first layer was ReLU activated and the next had a softmax activation to aid multi-class classification.

|       | Pre-Padded | | Post-Padded | |
|-------|------------|------|-------------|------|
| Model | Validation (%) | Test (%) | Validation (%) | Test (%) |
| **TCN** | 91.447 | **87.155** | 91.118 | **85.871** |
| BiLSTM | 90.013 | 83.669 | 89.144 | 81.467 |
| LSTM | 91.118 | 80.550 | 88.157 | 79.816 |
| FCNN | **91.776** | 79.266 | **92.763** | 81.100 |

Table 5: Comparison of deep learning models.

| Model | Trainable Parameters | Inference Time (s) [per 1500 samples] |
|-------|----------------------|---------------------------------------|
| **TCN** | 127k | **0.29** |
| BiLSTM | 134k | 0.45 |
| LSTM | **51.5k** | 0.36 |
| FCNN | 136k | 0.30 |

Table 6: The deep learning model parameters and inference time evaluated using a system with AMD Ryzen 5 3600 CPU and an NVIDIA RTX 3070 GPU with 8GB VRAM.

Prior works [18][23][24][25] have shown that TCN models tend to outperform LSTM models for time series data classification. This is also supported by the results obtained from the TCN model we have developed. In this model, all the convolutional layers are ReLU activated. Out of all the models developed, the TCN model generalized best over the test set, with pre and post-padded inputs. The model also exhibits lower variance as well as lower bias. It also has the lowest inference time out of all models developed, which can aid future real-time applications. The TCN model architecture is as illustrated in Fig. 4.

## 7. Results and discussions

The confusion matrix for the predictions made using the TCN model on the pre-padded test dataset is depicted in Fig. 5. Stroke-wise accuracies indicate that the model, at times, falsely predicts push strokes as topspin strokes and vice-versa. We attribute this to the fact that both these strokes are completed in a short duration, and have a low-arcing trajectory across the table. Few flat strokes are also predicted as topspin strokes, and this can be attributed to the high velocity of both strokes. Block and lob strokes have been classified with the highest accuracy due to the highly distinctive nature of the ball trajectories associated with them.

Our experiments with pre-padded and post-padded inputs show that models trained on pre-padded inputs outperform the models trained on post-padded inputs in terms of generalization. This aligns with the findings presented by Dwarampudi and Reddy [26].

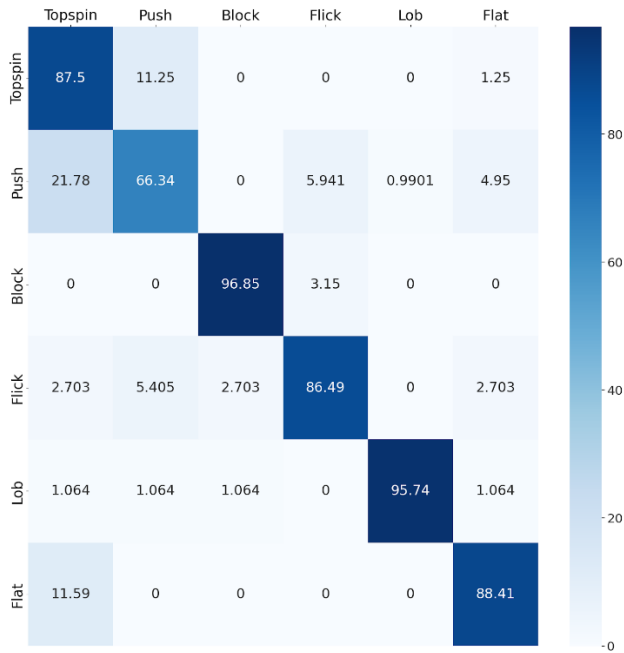

Figure 5: Confusion Matrix for the test results obtained using the TCN model.

## 8. Conclusion and future work

This paper describes a systematic method in which stroke detection and recognition is performed using ball trajectory data. The optimal camera position being the umpire's view of the table provides a complete view of the table, which helps us capture the entire trajectory of the stroke. Two distinct methods of ball-tracking have been compared and contrasted with, which helped in the labeling of the ball coordinates. This was necessary for the task of stroke detection and recognition. A mathematical model has been developed that utilizes the data from the ball tracking models to detect strokes and split the entire video into individual valid strokes. This data is then further used to train and validate the respective models for stroke recognition.

Expanding upon our current work, the use of multiple cameras can prove to be extensively resourceful. Three dimensional coordinates of the ball can be extracted using this method and can give us data that is crucial to analyze the spin of the ball. Information about the trajectory of the ball after the pitch is a very strong indicator of the amount of spin imparted on the ball and can help increase the accuracy of the stroke detection and recognition algorithms. Since these features are a crucial part of player and game analytics in table tennis, player profiling and game simulation merely using the information provided by the ball becomes a possibility. Additionally, the height of the ball from above the table and velocity of the ball before and after the pitch can be accurately determined by using pixel space measurements in tandem with depth and camera information. The information provided by these two characteristics of the ball is a good indicator of the quality of the stroke played. In particular, strokes played close to the height of the net, and certain strokes when played faster are more advantageous to the player. Along with the aforementioned attributes, the exact position in which the ball makes contact with the table can also be determined. This can be vastly useful for umpiring systems as well as helping players improve upon their performance.

Finally, viewers of the sport of table tennis find it difficult to follow the ball in a fast-paced rally on television. This has also caused changes to the size and material of the ball to make the sport more audience-friendly. Analytics derived from ball information when presented in a user-friendly manner can help make the viewing experience pleasant.